\pdfoutput=1

\documentclass[11pt]{article}

\usepackage[]{EMNLP2023}

\usepackage{times}
\usepackage{latexsym}

\usepackage[T1]{fontenc}

\usepackage[utf8]{inputenc}

\usepackage{microtype}

\usepackage{inconsolata}

\usepackage{amsmath}
\usepackage{amssymb}
\usepackage{amsthm}
\usepackage{amsfonts}
\usepackage{multicol}
\usepackage{multirow}
\usepackage{booktabs}
\usepackage{graphicx}
\usepackage{url}

\usepackage{color}
\definecolor{Red}{RGB}{255,0,0}
\definecolor{Green}{RGB}{0,176,80}
\definecolor{Blue}{RGB}{48,184,232}

\newcommand{\modelname}{SlotSum}
\newcommand{\datasetname}{WikiFactSum}

\title{Reducing Hallucinations in Entity Abstract Summarization with Facts-Template Decomposition}

\author{Fangwei Zhu, Peiyi Wang, Zhifang Sui \\
  National Key Laboratory for Multimedia Information Processing, Peking University \\
  \texttt{zhufangwei2022@stu.pku.edu.cn} \\
  \texttt{wangpeiyi9979@gmail.com} \\
  \texttt{szf@pku.edu.cn} \\
}

\begin{document}
\maketitle
\begin{abstract}
Entity abstract summarization aims to generate a coherent description of a given entity based on a set of relevant Internet documents.
Pretrained language models (PLMs) have achieved significant success in this task, but they may suffer from hallucinations, i.e. generating non-factual information about the entity.
To address this issue, we decompose the summary into two components: \textbf{Facts} that represent the factual information about the given entity, which PLMs are prone to fabricate; and \textbf{Template} that comprises generic content with designated slots for facts, which PLMs can generate competently.
Based on the facts-template decomposition, we propose \modelname{}, an explainable framework for entity abstract summarization.
\modelname{} first creates the template and then predicts the fact for each template slot based on the input documents. 
Benefiting from our facts-template decomposition, \modelname{} can easily locate errors and further rectify hallucinated predictions with external knowledge.
We construct a new dataset \datasetname{} to evaluate the performance of \modelname{}.
Experimental results demonstrate that \modelname{} could generate summaries that are significantly more factual with credible external knowledge.
\footnote{Code \& data available at \url{https://github.com/solitaryzero/SlotSum}}

\end{abstract}

\section{Introduction}

\begin{figure}[t]
    \centering
    \includegraphics[width=0.9\columnwidth]{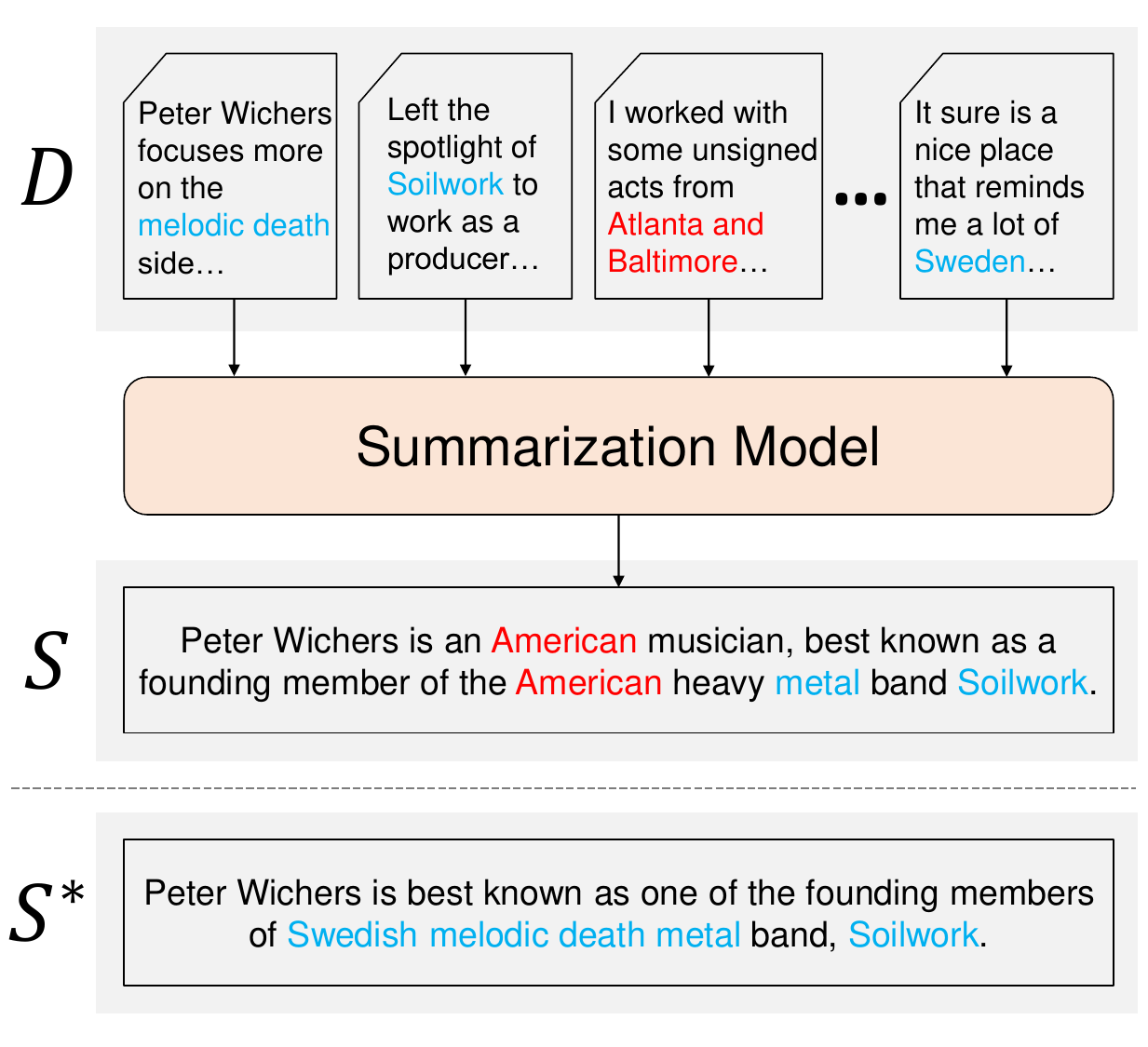}
    \caption{A demonstration of the entity abstract summarization task. The summarization model aims to generate a brief summary $S$ that resembles the golden summary $S^*$ about the person \textit{Peter Wichers} from source documents $D$ collected from the Internet. \textcolor{Blue}{Blue} indicates correct facts, and \textcolor{Red}{Red} indicates noise and hallucinated predictions.}
    \label{fig:intro_example}
\end{figure}

Entity abstract summarization aims to generate a correct and fluent summary of an entity based on a collection of relevant documents gathered from the Internet. 
The task is typically viewed as a variant of the multi-document summarization task, as illustrated in Figure \ref{fig:intro_example}.
Entity abstract summarization could provide users with condensed information about a given entity in search engines, and assists the writing process of new Wikipedia entries~\cite{sauper2009automatically}.

Early pieces of work~\cite{sauper2009automatically, banerjee2016wikiwrite} on entity abstract summarization generally adopt extractive methods that select valuable sentences from the input documents, which face the problem of poor fluency in the generated summary.
Inspired by the large-scale WikiSum~\cite{liu2018generating} dataset, recent work tends to directly generate abstractive summaries from the given input documents.
Various approaches have been explored to identify necessary information from the vast input documents: 
\citet{perez2019generating} classifies documents into different topics to filter out noisy data, HierSumm~\cite{liu2019hierarchical} ranks the input documents with a learning-based ranker, and models the dependency across different paragraphs, and NoisySumm~\cite{liu2021noisy} distills the knowledge in a noisy teacher model into a student model.

However, these models suffer from hallucinations as they are not able to ensure the factual correctness of generated summaries.
Hallucinations refer to the nonsensical or unfaithful contents in the generated texts~\cite{maynez2020faithfulness}.
They could hinder performance and introduce false information for real-world applications~\cite{ji2023survey}.
Take Figure \ref{fig:intro_example} as an example, the summarization model correctly identifies the band name and band genre of the given entity \textit{Peter Wichers}, but incorrectly assumes the nationality to be American rather than Sweden.
Hallucinated by the text snippet ``from Atlanta and Baltimore'' in the source, the summarization model mistakenly relates the text snippet to the nationality of America, ignoring the fact that \textit{Peter Wichers} comes from Sweden.
These factual errors are difficult to detect with automatic metrics and could degrade the system performance, leading to the failure in meeting user expectations in real-world scenarios~\cite{ji2023survey}.

Hallucination is an intrinsic drawback of language models and is challenging even for large language models~\cite{bubeck2023sparks}.
Hallucinations are difficult to eliminate under the traditional sequence-to-sequence paradigm, and various attempts have been taken to introduce external knowledge like dense retrieval~\cite{lewis2020retrieval} or API calling~\cite{schick2023toolformer}.
Templates are naturally suitable for utilising external knowledge, as the slots in templates could provide guidance for retrieving knowledge.
While template-based methods are widely adopted in various text generation problems like data-to-text\cite{luo2020make} and dialogue system~\cite{liang2021learning}, there has not yet been template-based solutions for entity abstract summarization.
To face the difficulty of locating hallucinations in plain text, we augment the language model to generate semi-structured templates before outputting the final summary.
We view the entity abstract as a combination of \textbf{Facts} and \textbf{Template}, where \textbf{Facts} represents the factual information about the given entity, and \textbf{Template} represents the remaining generic components.
In this way, we can easily locate the facts in generated summaries, and correct erroneous predictions with external knowledge like knowledge base or human assistance.

\begin{figure*}
    \centering
    \includegraphics[width=0.95\linewidth]{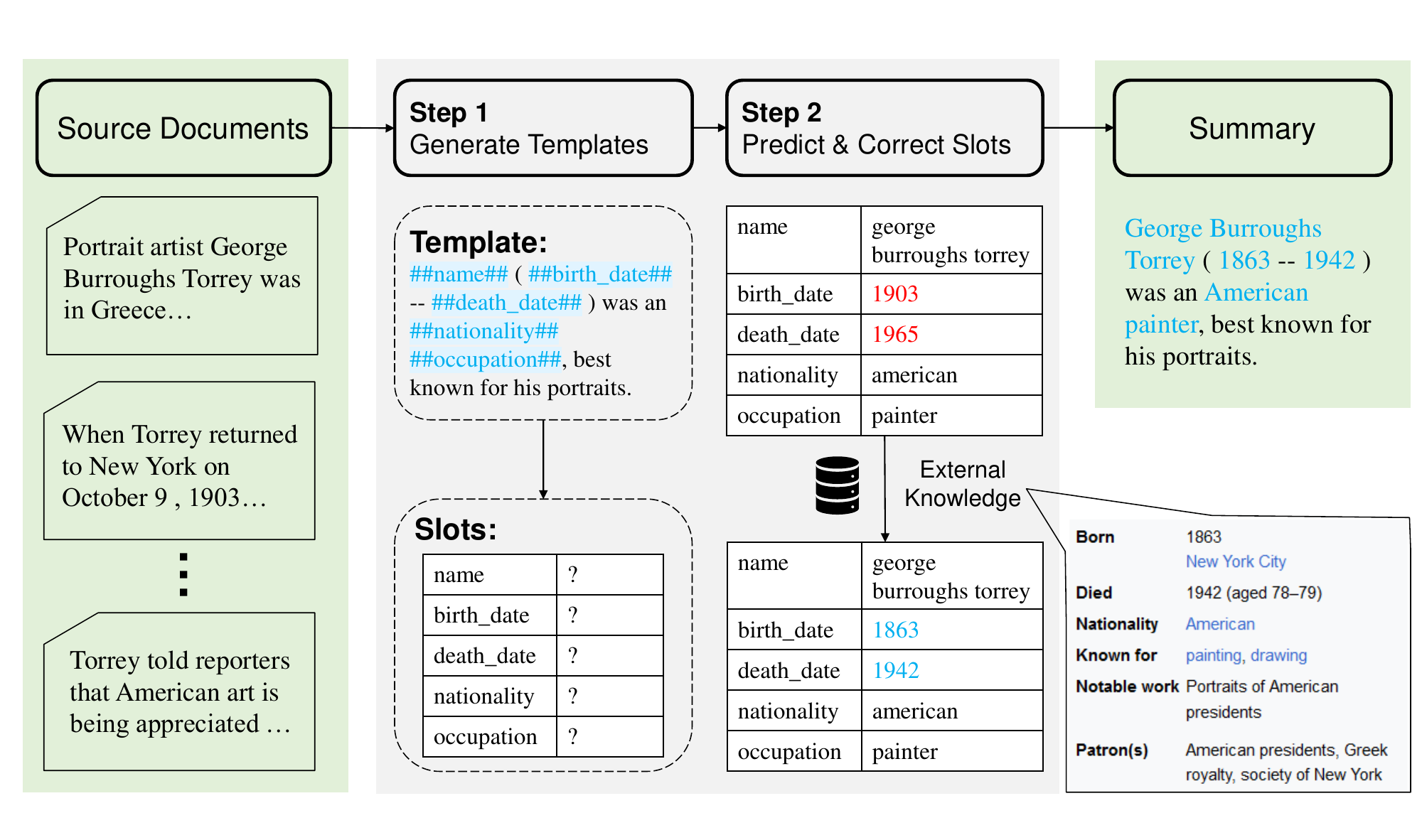}
    \caption{The overall framework of \modelname{}, with the entity \textit{George Burroughs Torrey} as an example. In step 1, \modelname{} generates a template from the source documents which contain several slots to be filled like \textit{name} and \textit{birth\_date}. In step 2, \modelname{} first predicts the slots according to the source documents and then corrects hallucinated predictions (like 1903 for \textit{birth\_date}) with credible external knowledge. Finally, The summary is constructed by filling slots into the template.}
    \label{fig:model_structure}
\end{figure*}

Based on the facts-template decomposition, in this paper, we propose \modelname{}, an explainable entity abstract summarization framework that disentangles facts from generated entity abstracts to reduce hallucination.
\modelname{} first generates a template according to the input documents, where the facts in the template are replaced by slots marked with special tokens.
Afterwards, \modelname{} predicts the facts for each slot, and attempts to rectify the predictions with credible external knowledge.
The final summary is constructed by filling facts into slots.
\modelname{} could ensure factual correctness by introducing external knowledge sources, while retaining the ability to generate summaries solely from the input documents.
Moreover, the framework is highly controllable and explainable, as the facts are explicitly stated and can be easily edited.

By supplementing the WikiSum~\cite{liu2018generating} dataset with related facts, we construct \datasetname{} to observe the effectiveness of disentangling facts.
\modelname, with BART-large~\cite{lewis2020bart} as the backbone model, clearly outperforms other baseline models on ROUGE score when credible external knowledge is provided.
We also evaluate the factual correctness with both automatic metrics and human evaluation, where \modelname{} significantly provides more correct facts.

In conclusion, the main contributions of our work are as follows:
\begin{itemize}
    \item We disentangle fact from the generated text in the entity abstract summarization task, which helps to reduce hallucinations produced by pretrained language models.
    \item We propose \modelname{}, an entity abstract summarization framework that first generates templates with slots, then predicts facts and further correct them with external knowledge to fill in the slots.
    \item We construct a new dataset \datasetname{}, and conduct a series of experiments on it. \modelname{} demonstrates a clear advantage over other baseline models, indicating disentangling facts from the text would lead to more factually correct summaries.
\end{itemize}

\section{Method}
In this section, we propose \modelname{}, an entity abstract summarization model that disentangles facts from the text to reduce hallucinations in generated summaries.
As illustrated in Figure \ref{fig:model_structure}, \modelname{} first trains a template generator module to generate templates from input documents, where the facts in the templates are replaced with slots to be filled.
Each slot is represented with a feature name that acts as the key of the fact.
Afterwards, \modelname{} predicts the slots with a slot predicting module that extracts relevant information from source documents given the feature name. 
As the predicted slots could contain hallucinations, we further correct the predictions with credible external knowledge sources.
In this way, \modelname{} is able to enhance the factual correctness with further external knowledge.

\subsection{Generating Templates}
\label{ssec:template_prediction}
Entity abstract summarization aims to generate an abstractive description $\mathcal{S} = (s_1, s_2, \ldots, s_m)$ about a given entity $e$ from a set of relevant text snippets $\mathcal{D} = (d_1, d_2, \ldots, d_n)$.
Given the dataset $\mathcal{C} = (\mathcal{D}, \mathcal{S})$, we first extract templates $\mathcal{T}$ from the golden summaries $\mathcal{S}$.
We denote the facts related to the entity $e$ with the form of feature pairs $\{(k, v)\}$, where the key of each pair $k$ represents the feature name and the value $v$ represents the fact content. 
Each fact $v$ about the given entity in the original text $s$ are replaced with a slot $t$, and each slot $t$ is represented by its feature name $k$:
\begin{equation}
    t(v) = \texttt{[SLT]}\ k\ \texttt{[/SLT]}
\end{equation}
where $\texttt{[SLT]}$ and $\texttt{[/SLT]}$ are special tokens.

To gather the training data for the template generation model, we construct golden templates by replacing the facts in the golden summary with slots according to external fact pairs $\{(k^*, v^*)\}$. 
Notice that the fact value $v^*$ does not necessarily have an exact match in the original text $s$, we adopt the fuzzy matching technique to discover more potential slots.
For each fact content $v^*$, we search for a substring $s_{v}$ in $s$ which has the highest score $sim(v^*, s_{v})$ under a given similarity metric $sim$.
We replace $s_{v}$ with a slot when its similarity score $sim(v^*, s_{v})$ surpasses a threshold $\delta$, and keep it unchanged otherwise:
\begin{equation}
    s_{v} \gets \left\{
    \begin{aligned}
        &t(v^*),\ sim(v^*, s_{v}) \geq \delta \\
        &s_{v},\ \text{otherwise}
    \end{aligned}
    \right.
\end{equation}
where the threshold $\delta$ is a hyperparameter.
In this way, we replace the facts in $\mathcal{S}$ with slots and obtain the knowledge-agnostic template set $\mathcal{T}$.

Predicting templates can be viewed as a standard sequence-to-sequence task, where the input is web documents $\mathcal{D}$ and the output is the desired templates $\mathcal{T}$.
As the slots are represented by feature keys $k$ rather than values $v$, the model does not have access to external factual knowledge in the training process, and therefore will not cause data leakage.

We use BART as the backbone of the template predictor and train the model with cross-entropy loss.
The normalized sorted Indel distance is selected as the string similarity metric $sim$:
given two strings $a$ and $b$, we first sort the tokens in the strings, and then calculate the Levenshtein distance $L(a,b)$ between $a$ and $b$ where only insertions and deletions are allowed.
The distance is normalized as the final similarity:
\begin{equation}
    sim(a,b) = 1 - \frac{L(a,b)}{Len(a)+Len(b)}
\end{equation}

\subsection{Predicting and Correcting Slots}
After generating the template, we need to fill facts into the template slots to obtain the final summary.
\modelname{} first predicts the facts according to input documents, then corrects the facts with credible external knowledge.

\subsubsection{Predict from Input Documents}
Firstly, \modelname{} tries to extract facts from input documents to fill in the slots. 
We use a simple sequence-to-sequence slot predictor to extract desired facts from the source documents, whose input can be denoted as:
\begin{equation}
    \texttt{[CLS]}\ n\ k\ \texttt{[SEP]}\ d\ \texttt{[SEP]}
\end{equation}
where $n$ is the name of the entity to be summarized, $k$ is the slot name, $d$ refers to the source documents, $\texttt{[CLS]}$ and $\texttt{[SEP]}$ are special tokens.
The slot predictor is trained on the golden fact set with the label smoothed cross-entropy loss.
We denote the predicted results as $\{(k_p, v_p)\}$.

\subsubsection{Correct with External Knowledge}
By introducing external knowledge sources, \modelname{} is able to ensure the correctness of facts.
An external knowledge source, for example, knowledge bases and human experts, will provide a set of trustworthy facts, which could fill the slots without producing hallucination.
Similar to the slot replacing scenario in Section \ref{ssec:template_prediction}, the slot names do not exactly match the fact feature names.
Therefore, we only correct slots that have a candidate with name similarity greater than a threshold $\delta$.
We denote the corrected results as $\{(k_c, v_c)\}$, where the key set $K_c$ contains corrected slots.

\subsubsection{Slot-Filling Strategies}
\label{sssec:slot_strategies}
We adopt three distinct slot-filling strategies to get the slot value $v$ for each slot $k$:

\paragraph{\textbf{Discard}} 
The discard strategy only fills slots corrected with external knowledge, and fills other slots with empty strings to ensure the precision of facts.
This strategy can be formulated as:
\begin{equation}
    v = \left\{
    \begin{aligned}
        v_c &, k \in K_c, \\
        \varnothing &, k \notin K_c.
    \end{aligned}
    \right.
\end{equation}

\paragraph{\textbf{Predict}} 
The predict strategy predicts all slots first and then tries to correct the slots with external knowledge, which maximizes the information recall in the generated summary.
This strategy can be formulated as:
\begin{equation}
    v = \left\{
    \begin{aligned}
        v_c &, k \in K_c, \\
        v_p &, k \notin K_c.
    \end{aligned}
    \right.
\end{equation}

\paragraph{\textbf{All-Predict}}
The all-predict strategy predicts all slots without utilizing external knowledge, which is applicable when no additional knowledge is provided.
This strategy can be formulated as:
\begin{equation}
    v = \left\{
    \begin{aligned}
        v_p &, k \in K_c, \\
        v_p &, k \notin K_c.
    \end{aligned}
    \right.
\end{equation}

\begin{table}[htbp]
    \centering
    \begin{tabular}{c|c}
    \toprule
    Feature & Value \\
    \midrule
    \# Examples & 27372 \\
    Train/Valid/Test & 21934/2697/2741 \\
    \# Slots & 83782 \\
    Avg. Slots & 3.06 \\
    \# Keys & 367642 \\
    Avg. Keys & 13.43 \\
    Avg. Value Len & 4.11 \\
    Avg. Src Length & 2000 \\
    Avg. Tgt Length & 26.47 \\
    \bottomrule
    \end{tabular}
    \caption{Statistics of \datasetname{}.}
    \label{tab:dataset_stats}
\end{table}

\section{Experiments}
\subsection{Experimental Setup}
\paragraph{Dataset Construction}

\begin{table*}[htbp]
    \centering
    \begin{tabular}{l|c c c|c c c|c|c}
        \toprule
        \multirow{2}{*}{Model} & \multicolumn{3}{c|}{ROUGE} & \multicolumn{3}{c|}{BERTScore} & FactCC & Unieval\\
         & 1 & 2 & L & P & R & F & & \\
        \midrule
        BART-large & 45.96 & 26.30 & 43.69 & 0.911 & 0.871 & 0.890 & 0.416 & 0.203 \\
        T5-base & 40.44 & 21.29 & 38.67 & 0.893 & 0.870 & 0.881 & 0.218 & 0.117 \\
        PEGASUS-xsum & 38.67 & 20.57 & 37.15 & 0.867 & 0.809 & 0.836 & 0.454 & 0.217 \\
        NoisySumm & 43.44 & 23.96 & 41.19 & 0.901 & 0.879 & 0.889 & 0.326 & 0.171 \\
        \midrule
        \modelname{} (Discard) & \textbf{59.91} & \textbf{38.53} & \textbf{56.30} & \textbf{0.929} & 0.895 & 0.911 & \textbf{0.585} & \textbf{0.258} \\
        \modelname{} (Predict) & 59.53 & 38.17 & 55.96 & 0.927 & \textbf{0.901} & \textbf{0.914} & 0.559 & 0.222 \\
        \modelname{} (All-Predict) & 46.14 & 26.12 & 44.35 & 0.915 & 0.884 & 0.899 & 0.207 & 0.095 \\
        \bottomrule
    \end{tabular}
    \caption{Automatic evaluation results on \datasetname{}. \textbf{Bold} numbers indicate the best performance on each metric. \modelname{} refers to our proposed method, and Discard, Predict and All-Predict refer to the 3 slot-filling strategies in Section \ref{sssec:slot_strategies} respectively.}
    \label{tab:automatic_results}
\end{table*}

We conduct our experiments on \datasetname{}, a proposed dataset constructed by combining WikiSum~\cite{liu2018generating} and WikiBio~\cite{lebret2016neural}.
WikiSum is a widely-used entity abstract summarization dataset, which aims to generate the Wikipedia abstract of an entity from websites searched by Google.
WikiBio provides structured data about a specific person and aims to generate a biographical abstract about that person. 
WikiSum contains only textual data, thus introducing WikiBio could complement WikiSum with facts.

In the WikiBio dataset, we select people with two occupations to construct our \datasetname{}: \textit{Artist} and \textit{Soccer Player}, whose features are distinct to each other.
The entries with selected occupations are extracted from the WikiAsp~\cite{hayashi2021wikiasp} dataset, which is also derived from WikiSum. 
As WikiAsp (and the original WikiSum dataset) does not provide the Wikipedia title of each example, we try to find the corresponding entries in WikiBio by fuzzy matching.
We compute the bag-of-words similarity $S_{bow}$ of the abstract of entry $a$ in WikiAsp and the abstract of entry $b$ in WikiBio:
\begin{equation}
    S_{bow} = \frac{A \cap B}{A \cup B}
\end{equation}
where $A$ and $B$ are the word set of $a$ and $b$.
We assume $a$ and $b$ are identical if $S_{bow} > 0.8$, and map the corresponding entry between WikiAsp (which is a subset of WikiSum) and WikiBio.

By combining the fact pairs $F = \{(k, v)\}$ in WikiBio with the source document $D$ and the desired summary $S$ in WikiSum, we construct our proposed dataset \datasetname{}.
In this way, each entry in \datasetname{} consists of three parts: the source document $D$, the desired summary $S$, and external facts $F = \{(k, v)\}$.
Table \ref{tab:dataset_stats} demonstrates the detailed parameters of \datasetname{}.

\paragraph{Baselines and Metrics}
We compare our model with three pretrained autoregressive language models BART~\cite{lewis2020bart}, T5~\cite{raffel2020exploring}, and PEGASUS~\cite{zhang2020pegasus}, which are widely used in summarization tasks.
We also choose NoisySumm~\cite{liu2021noisy} as a baseline model, which is an entity abstract summarization model that filters noise information by distilling the teacher model into a student model.
Four metrics are adopted to evaluate the performance of models: 
\begin{itemize}
    \item The widely used token-level similarity metric ROUGE~\cite{lin2004rouge};
    \item BERTScore~\cite{zhang2019bertscore} that evaluates text similarity with BERT;
    \item FactCC~\cite{kryscinski2020evaluating} that identifies conflicts between source documents and claims; \item UniEval~\cite{zhong2022towards} that learns multi-dimensional evaluators from intermediate tasks.
\end{itemize}

\paragraph{Implementation Details}
We choose BART-large~\cite{lewis2020bart} as the backbone model of \modelname{}, and set the similarity threshold $\delta = 0.8$.
The model is trained on a single NVIDIA RTX 3090 for 4 epochs with a learning rate of 1e-5.
Specifically, we use T5-base and PEGASUS-xsum as baseline models, which match the scale of BART-large.

\begin{table*}[htbp]
    \centering
    \begin{tabular}{l|c c c c|c}
        \toprule
        \multirow{2}{*}{Model} & \multicolumn{4}{c|}{Linguistic Quality} & \multirow{2}{*}{QA} \\
        & C & F & S & Overall & \\
        \midrule
        BART-large & \textbf{5.00} & \textbf{5.00} & 4.87 & \textbf{4.96} & 1.458 \\
        T5-base & 4.57 & 4.73 & 4.50 & 4.62 & 1.208 \\
        PEGASUS-xsum & 4.27 & 4.80 & 4.80 & 4.66 & 1.192 \\
        NoisySumm & 4.77 & 4.80 & 4.80 & 4.91 & 1.205 \\
        \midrule
        \modelname{} (Discard) & 4.33 & 4.93 & \textbf{4.90} & 4.79 & 2.048 \\
        \modelname{} (Predict) & 4.80 & \textbf{5.00} & \textbf{4.90} & 4.90 & \textbf{2.075} \\
        \modelname{} (All-Predict) & 4.73 & \textbf{5.00} & \textbf{4.90} & 4.88 & 1.460 \\
        \bottomrule
    \end{tabular}
    \caption{Human evaluation results on \datasetname{}. \textbf{Bold} numbers indicate the best performance on each metric. C, F, and S represent completeness, fluency, and succinctness respectively.}
    \label{tab:human_results}
\end{table*}

\subsection{Main Experiment Results}
\label{ssec:main_exp}
Table \ref{tab:automatic_results} demonstrates the automatic evaluation results on \datasetname{}, from which several insights could be concluded:
\paragraph{Credible Knowledge Matters.} 
After correcting the summaries with external knowledge, it can be observed that \modelname{} shows a clear advantage over other baseline models on both metrics.
We can safely assume that introducing external knowledge will drastically improve the quality of generated summaries.
\paragraph{General Summarization Models Struggles.}
T5 and PEGASUS have been specially pretrained on summarization tasks, however, they perform poorly on entity abstract summarization, even worse than BART.
This phenomenon indicate that the abilities required for entity abstract summarization are not identical to those required for general summarization tasks.
The input documents of entity abstract summarization contain much more noise, which would mislead general summarization models to low-quality summaries.
\paragraph{Unchecked Predictions are Dangerous.}
Compared with baseline models, \modelname{} with the all-predict strategy seems to be performing slightly better on ROUGE and BERTScore;
However, on fact-oriented metrics FactCC and Unieval, the all-predict strategy obviously falls behind, performing even worse than the baseline BART model.
We infer that templates tend to guide the model to discover more facts in order to fill in the slots, whereas the slot predicting module may fabricate facts, in turn leading to the detriment of factual correctness.
In practice, it should be paid extra caution not to completely trust the predictions.
Subsequent manual checking may be very beneficial to the final summary.

\begin{table*}[htbp]
    \centering
    \begin{tabular}{l|c c c|c c c|c|c}
        \toprule
        \multirow{2}{*}{Model} & \multicolumn{3}{c|}{ROUGE} & \multicolumn{3}{c|}{BERTScore} & FactCC & Unieval \\
         & 1 & 2 & L & P & R & F & & \\
        \midrule
        BART-large+K & 48.05 & 27.89 & 45.68 & 0.912 & 0.888 & 0.899 & 0.304 & 0.169 \\
        T5-base+K & 43.46 & 23.80 & 41.46 & 0.902 & 0.877 & 0.889 & 0.211 & 0.127  \\
        PEGASUS-xsum+K & 41.64 & 22.79 & 39.92 & 0.886 & 0.844 & 0.864 & 0.376 & 0.193  \\
        \midrule
        \modelname{} (K+Discard) & \textbf{62.55} & \textbf{41.30} & \textbf{58.70} & 0.934 & 0.906 & 0.920 & \textbf{0.592} & \textbf{0.263} \\
        \modelname{} (K+Predict) & 62.46 & 41.23 & 58.58 & \textbf{0.934} & \textbf{0.907} & \textbf{0.920} & 0.577 & 0.256 \\
        \modelname{} (K+All-Predict) & 48.03 & 27.87 & 46.08 & 0.920 & 0.886 & 0.903 & 0.213 & 0.108 \\
        \midrule
        BART-large+KV & \textbf{69.88} & \textbf{54.13} & \textbf{66.68} & \textbf{0.949} & \textbf{0.923} & \textbf{0.935} & \textbf{0.842} & \textbf{0.360} \\
        T5-base+KV & 67.66 & 51.11 & 64.51 & 0.945 & 0.916 & 0.930 & 0.831 & 0.321 \\
        PEGASUS-xsum+KV & 63.10 & 43.06 & 60.30 & 0.936 & 0.900 & 0.917 & 0.753 & 0.185 \\
        \midrule
        \modelname{} (KV+Discard) & 68.08 & 48.09 & 63.99 & 0.942 & 0.917 & 0.929 & 0.787 & 0.354 \\
        \modelname{} (KV+Predict) & 68.20 & 48.17 & 64.06 & 0.942 & 0.917 & 0.929 & 0.773 & 0.349 \\
        \modelname{} (KV+All-Predict) & 53.81 & 34.43 & 51.49 & 0.921 & 0.897 & 0.910 & 0.243 & 0.122 \\
        \bottomrule
    \end{tabular}
    \caption{Ablation results on \datasetname{}. \textbf{K} indicates taking the keys as part of the input, while \textbf{KV} indicates taking the full key-value pair list as part of the input. \textbf{Bold} numbers indicate the best performance on each metric under certain setting (K or KV). \modelname{} refers to our proposed method, and Discard, Predict and All-Predict refer to the 3 slot-filling strategies in Section \ref{sssec:slot_strategies} respectively.}
    \label{tab:ablation_results}
\end{table*}

\subsection{Human Evaluation}
We conduct human evaluation consisting of two parts on the performance of different models.
Each part is annotated by 3 annotators on 20 examples randomly sampled from the test set.

The first part evaluates the linguistic quality of generated summaries.
We require the annotators to evaluate the linguistic quality from three aspects: \textbf{Completeness} (is the summary informative?), \textbf{Fluency} (is the summary grammatically correct?), and \textbf{Succinctness} (does the summary get rid of redundant text?).
The score of each metric ranges from 1 to 5, and the final result average is obtained by averaging the score of three annotators.

The second part evaluates the factual correctness of generated summaries, where we follow the question-answering (QA) scheme~\cite{clarke2010discourse} for examination.
We manually create 3 questions on the base of golden summaries for each example, and ask annotators to answer these questions by viewing generated summaries as backgrounds.
An expert then examines the annotated answers and assigns scores to different models.
A model gets 1 point for a correct answer and 0 point for a wrong answer.
Specially, we give 0.1 point for a model if a question is not answerable on the base of generated summary, as it does not cause hallucination.

From Table \ref{tab:human_results}, we can observe that all models achieve decent scores on linguistic quality metrics, while BART-large performs the best.
The speculation why \modelname{} gets a lower score on \textbf{Completeness} is that some information cannot be conveyed through a single slot, which leads to minor information loss compared with the original BART.
On the scope of QA examination, the factual correctness of generated summaries significantly increased after calibrating slot values with external knowledge, which proves the effectiveness of introducing external knowledge from another perspective.

\subsection{Ablation Study}
To further investigate how facts influence the final summary, we evaluate the models\footnote{NoisySumm does not fit the data augmentation, and is thus excluded from ablation study.} under two alternative settings: (1) Taking the keys $\{k\}$ as part of the input, denoted as +K; (2) Taking the full key-value pair list $\{(k, v)\}$ as part of the input, denoted as +KV. 

The keys or key-value pairs are serialized by concatenating them with special tokens $\texttt{[KV1]}$ and $\texttt{[KV2]}$:
\begin{equation}
    k_1\ \texttt{[KV1]}\ k_2\ \ldots
\end{equation}
\begin{equation}
    k_1\ \texttt{[KV1]}\ v_1\ \texttt{[KV2]}\ k_2\ \texttt{[KV1]}\ v_2 \ldots
\end{equation}
where we use the token '|' for $\texttt{[KV1]}$ and '\#' for $\texttt{[KV2]}$ in practice.
The serialized string $f_s$ is then concatenated with the source document $d$ with the separation token $\texttt{[SEP]}$ as the final input:
\begin{equation}
    \texttt{[CLS]}\ f_s \ \texttt{[SEP]}\ d\ \texttt{[SEP]}
\end{equation}

Table \ref{tab:ablation_results} demonstrates the results under different settings, from which we draw three conclusions:

\paragraph{\modelname{} Maintains Its Competitiveness.}
As expected, \modelname{} with the predict or discard strategy still maintains a clear advantage over other models on all metrics under the +K setting, demonstrating its ability in generating factually correct entity summaries.
\paragraph{Guidance from Keys Degrades Baseline Performance.}
An interesting discovery is that under the +K setting, baseline models guided by keys gain a boost on ROUGE and BERTScore, but perform worse on FactCC and Unieval instead.
This tendency is identical with the main experiment performance of \modelname{} under the all-predict strategy, confirming the hypothesis in section \ref{ssec:main_exp} that guiding the model with facts needed would in turn induce hallucination and harm its factual correctness.
\paragraph{Providing Golden Facts Wipes the Gap.}
We observe that when adding the full key-value pairs to the input, the scores achieved by \modelname{} are slightly lower than those achieved by BART, and the gaps between baseline models are also narrower.
Considering that the entity abstract summarization problem is essentially degenerated into the data-to-text problem under the +KV setting, we think it is understandable as direct filling values into slots inevitably affects the fluency of summaries.

\begin{figure*}
    \centering
    \includegraphics[width=\linewidth]{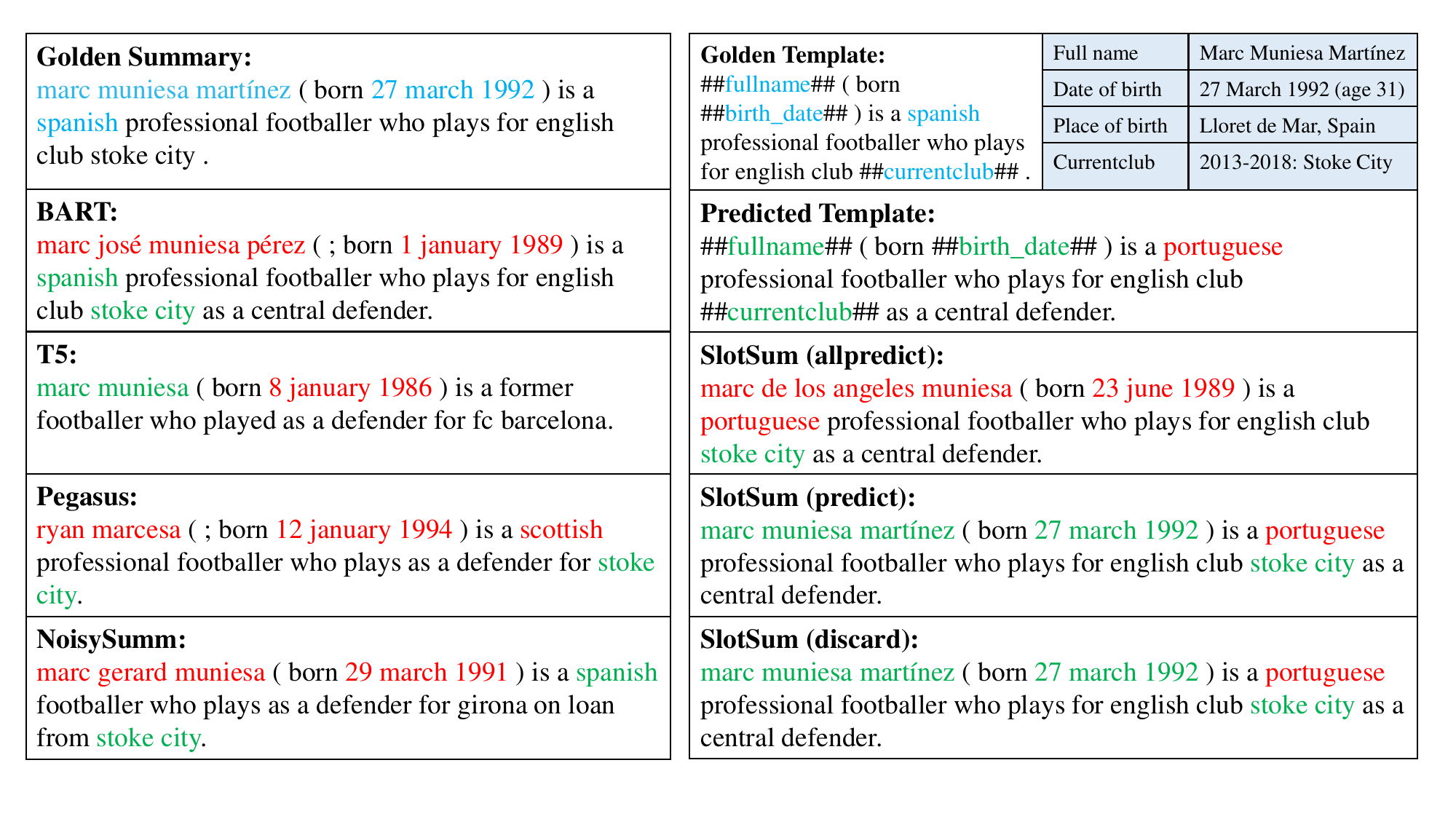}
    \caption{A case study on the entity \textit{Marc Muniesa Martínez}. \textcolor{Blue}{Blue} texts indicate golden facts, \textcolor{Green}{Green} texts indicate correct predictions, and \textcolor{Red}{Red} texts indicate hallucinated predictions.}
    \label{fig:case_study}
\end{figure*}

\subsection{Case Study}
\label{sec:case_study}
Figure \ref{fig:case_study} demonstrates the summaries generated by different models about the entity \textit{Marc Muniesa Martínez}.
We found two major observations by comparing these summaries:
\paragraph{External Knowledge is Crucial to Improving Factual Correctness.}
Factual errors occur across different models, for example, it can be clearly observed that every model fails in predicting the exact birth date of the person.
The factual errors are difficult to detect and correct, but can be easily mitigated by filling external knowledge into slots.
\paragraph{Hallucinated Facts Still Lurk in Templates.} 
Despite efficiently reducing hallucination, \modelname{} still fails to disentangle all facts from the template.
The nationality of \textit{Marc Muniesa Martínez} is not explicitly stored in the external fact list, and \modelname{} generates the hallucinated nationality \textit{Portuguese} as a part of the template rather than disentangling it as a slot.

To sum up, by disentangling facts from templates and introducing external knowledge, \modelname{} improves the factual correctness in an explainable manner, and provides guidance on which facts are more important.
However, we should be cautious that the templates cannot completely get rid of the hallucinations, and further check on factual correctness is still necessary.

\section{Related Work}

\subsection{Entity Abstract Summarization}
Entity abstract summarization aims to generate a concise description of a given entity from relevant texts gathered from the internet, and is generally viewed as a variant of multi-document summarization problem.
Compared with common multi-document summarization tasks, the input documents of entity abstract summarization contain more noise, which will hallucinate the result.

Early pieces of work mainly take an extractive approach.
\citet{sauper2009automatically} first delves into the entity abstract summarization problem, which selects useful sentences with Integer Linear Programming (ILP).
WikiWrite~\cite{banerjee2016wikiwrite} further takes the coherence scores between selected sentences into consideration to improve the linguistic quality of the final summary.

With the development of transformer models, WikiSum~\cite{liu2018generating} proposes an extractive-then-abstractive framework.
The framework first selects salient paragraphs according to TF-IDF scores, and then summarizes the selected paragraphs with a transformer decoder.
Meanwhile, WikiSum constructs a large-scale dataset with the same name, which gathers webpages relevant to the given entity via search engines as the input document, and views the abstract section of Wikipedia entries as golden outputs.

Inspired by WikiSum, succeeding models improve the quality of generated summaries from various aspects.
WikiCatSum~\cite{perez2019generating} utilizes the Latent Dirichlet Allocation (LDA) model to render the decoder topic-aware,
and HierSumm~\cite{liu2019hierarchical} adopts a learning-based model for the extractive stage, and computes the attention between paragraphs to model the inter-paragraph dependencies.
Multiple attempts have also been made to improve the factual correctness: TWAG~\cite{zhu2021twag} divides input documents into different topics to capture different aspects of facts, and NoisySumm~\cite{liu2021noisy} distills the base model into a student model to make it more robust to noise.
However, these models still suffer from hallucinations and produce non-factual but seemingly plausible summaries.

\subsection{Template-based Text Generation}
In text generation problems, using templates could give extra control to the generation process, and improve the readability and stability of output text.
Thus, template-based techniques are widely adopted in various fields like dialog generation, data-to-text, and even summarization.
Different tasks present different challenges, and the template-based methods are motivated to design specialized template decomposition frameworks for each unique task.
For example, the data-to-text model TS2~\cite{luo2020make} first constructs slots with provided data, and then stitches the slots with additional text to form a fluent sentence;
the task-oriented dialogue system NTRD~\cite{liang2021learning} generates a template with only one special slot for recommending items, which acts as textual context to improve the precision of item selection;
the summarization model $\text{Re}^3\text{Sum}$~\cite{cao2018retrieve} retrieves similar summaries in the training dataset as soft templates, improving the readability and stability of seq2seq summarization systems.
To the best of our knowledge, our work is the first to develop a template-based generation method for entity abstract summarization.
Specifically, we design a facts-template decomposition framework that explicitly integrates external knowledge to reduce the hallucinations in entity abstract summarization.

\section{Conclusion}
In this paper, we propose \modelname{}, a framework for entity abstract summarization that reduces hallucinations by disentangling facts.
\modelname{} views the summary of an entity as a combination of facts and fact-agnostic template, and generates the template first.
To fill the slots in the template, \modelname{} predicts facts from the input documents, and further corrects the erroneous predictions with external knowledge sources.
In this way, the hallucinations in generated summaries could be reduced in an explainable and controllable way.

We combine WikiSum~\cite{liu2018generating} with WikiBio~\cite{lebret2016neural} to construct \datasetname{}, an entity abstract summarization dataset which provides trustworthy knowledge aside from the input documents.
A series of experiments are conducted on \datasetname{} to compare the performance of \modelname{} against various baseline models.
Both automatic and human evaluation results prove that \modelname{} effectively reduces hallucinations after correcting slots with external knowledge in an explainable manner, while retaining comparable performance without additional knowledge against the baseline models.
In the future, we will further explore the fusion between knowledge and large language models, and explore a better approach to explicitly incorporating knowledge.

\section*{Limitations}
Our work is still limited in some aspects.
First, our proposed dataset is limited to the domain of biographies.
To generalize our model to broader domains, more training data is needed.

Second, we do not test the performance of large language models like ChatGPT, as the large language models may memorize the golden summaries (see Appendix \ref{app:leak}).
A more complete and more novel dataset is required to study the framework on LLMs.

\paragraph{Potential Risks.}
Our proposed dataset \datasetname{} uses WikiSum as the data source, which is built on a frozen version of Wikipedia.
The contents in Wikipedia will change along with time, and \datasetname{} may contain incorrect information. 

\section*{Ethics Statement}
In this section, we will discuss the ethical considerations of our work.
\paragraph{Licenses and terms.}
WikiBio is shared under the CC BY-SA 3.0 license, and WikiAsp is shared under the CC BY-SA 4.0 license.
These datasets are used in entity abstract summarization by multiple pieces of research, and we believe that they have been anonymized and desensitized.
\paragraph{Human Annotation.}
We recruited 3 human annotators to annotate the linguistic qualities and answer the questions, and 1 expert annotator to determine the golden answer for questions and score the answers of the 3 annotators.
These annotators are employed by commercial data annotation companies.
These recruited annotators have been paid with mutually agreed rewards under the agreed working time and price.
The annotators have been informed about the usage of annotated data, which has been recorded in the contract.
\paragraph{Intended use.}
Researchers are intended to use \datasetname{} to investigate the performance of entity abstract summarization with trustworthy knowledge.
WikiBio and WikiSum are intended for summarization research, which is compatible with our work.

\section*{Acknowledgements}

\bibliography{anthology,custom}
\bibliographystyle{acl_natbib}

\appendix

\begin{table}[htbp]
    \centering
    \begin{tabular}{c|c|c}
        \toprule
        Slot & Frequency & Popularity \\
        \midrule
        name & 21743 & 79.44\% \\
        birth\_date & 21595 & 78.89\% \\
        position & 5041 & 18.42\% \\
        currentclub & 4817 & 17.6\% \\
        occupation & 4595 & 16.79\% \\
        fullname & 3712 & 13.56\% \\
        death\_date & 3286 & 12.0\% \\
        birth\_place & 3089 & 11.29\% \\
        nationality & 2696 & 9.85\% \\
        genre & 2338 & 8.54\% \\
        \bottomrule
    \end{tabular}
    \caption{The 10 most frequent slots in \datasetname{}.}
    \label{tab:top_slots}
\end{table}

\begin{figure*}
    \centering
    \includegraphics[width=\linewidth]{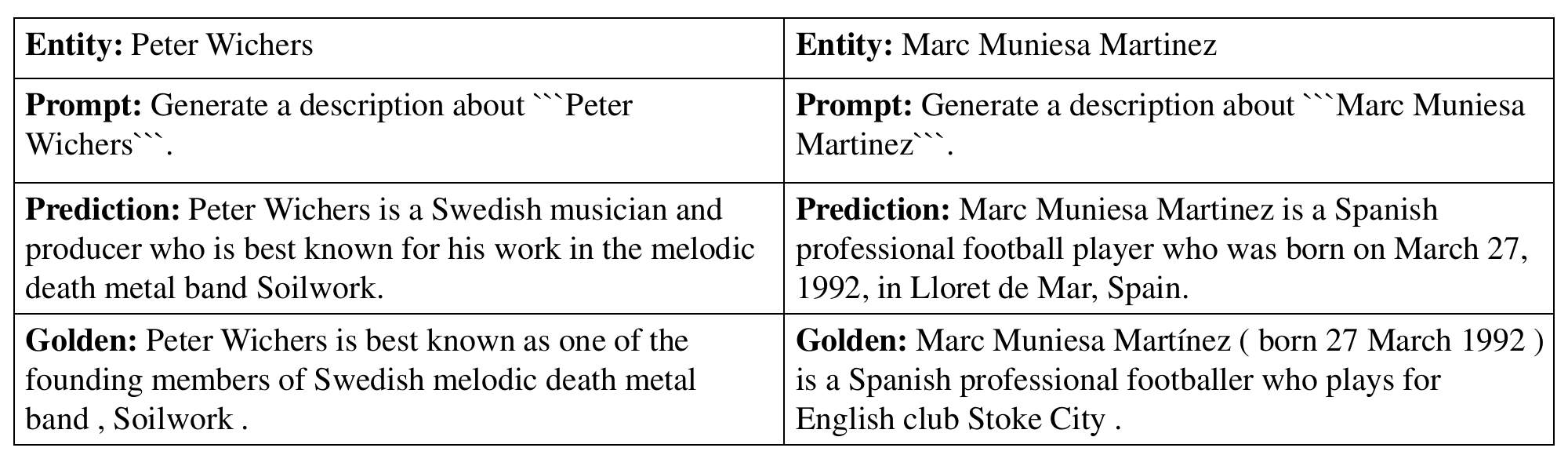}
    \caption{Data leakage in ChatGPT, with the entity \textit{Peter Wichers} and \textit{Marc Muniesa Martinez} as examples.}
    \label{fig:data_leak}
\end{figure*}

\begin{figure}
    \centering
    \includegraphics[width=\columnwidth]{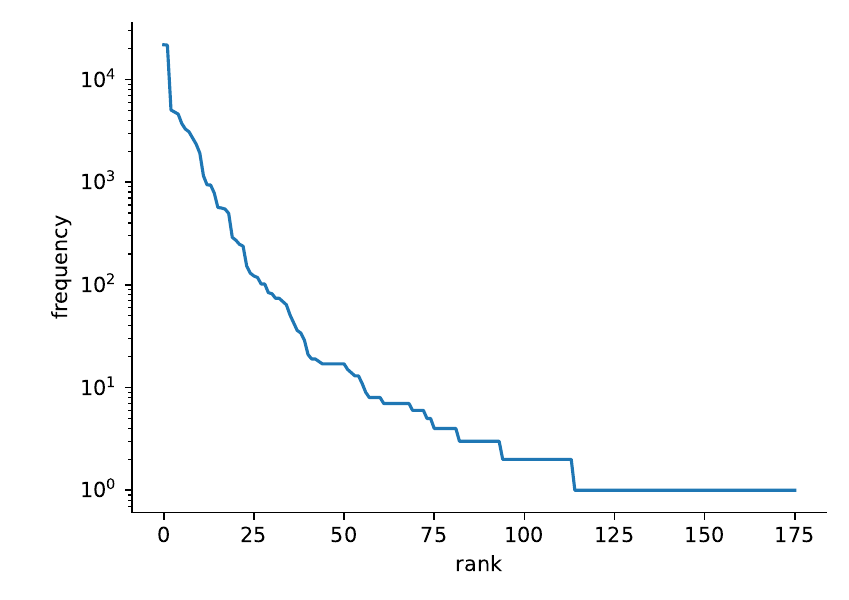}
    \caption{The frequency of different slots in \datasetname{}.}
    \label{fig:slot_freq}
\end{figure}

\section{Data Leakage in LLMs}
\label{app:leak}
We do not take large language models like ChatGPT as baseline models, as we notice obvious data leakage in large language models.
Take Figure \ref{fig:data_leak} as an example, by prompting ChatGPT with simple instructions, the model could generate summaries similar to the golden summary even without input documents.
A possible explanation is that LLMs remember the Wikipedia summaries, which act as the golden answers in WikiSum, in the pretraining process.
As a result, LLMs could simply ``recall'' these memories when queried.
We think it is unfair to compare other models with LLMs due to data leakage, and we do not take LLMs as baseline models.

\section{Slot Distribution In \datasetname{}}
Figure \ref{fig:slot_freq} shows the frequency of slots in \datasetname{}, and Table \ref{tab:top_slots} lists 10 most frequent slot names in \datasetname{}.
It can be observed that the frequency of slots has a long-tailed distribution.

\end{document}